\documentclass[letterpaper]{article} % DO NOT CHANGE THIS
\usepackage{arxiv2027}  % DO NOT CHANGE THIS
\usepackage[hyphens]{url}  % DO NOT CHANGE THIS
\usepackage{graphicx} % DO NOT CHANGE THIS
\usepackage{natbib}  % DO NOT CHANGE THIS AND DO NOT ADD ANY OPTIONS TO IT
\usepackage{caption} % DO NOT CHANGE THIS AND DO NOT ADD ANY OPTIONS TO IT
\usepackage{algorithm}
\usepackage{algorithmic}
\usepackage{multirow}
\usepackage{amsmath}
\usepackage{amssymb}
\usepackage{newfloat}
\usepackage{listings}
\DeclareCaptionStyle{ruled}{labelfont=normalfont,labelsep=colon,strut=off} % DO NOT CHANGE THIS
\floatstyle{ruled}
\newfloat{listing}{tb}{lst}{}
\floatname{listing}{Listing}

\usepackage{booktabs}

\title{SBMVTrack: Spike-Budgeted Multi-View Learning for Power-Efficient \\ UAV Tracking}
\author{
    Pengzhi Zhong\textsuperscript{\rm 1},
    Jiwei Mo\textsuperscript{\rm 2},
    Haolun Li\textsuperscript{\rm 1},
    Ge Zheng\textsuperscript{\rm 1},
    Jingqi Wang\textsuperscript{\rm 1},
    Xinyi Bo\textsuperscript{\rm 1},
    Shuiwang Li\textsuperscript{\rm 1}\corresponding
}

\affiliations{
    \textsuperscript{\rm 1}College of Computer Science and Engineering, 
    Guilin University of Technology, Guilin 541004, China\\
    \textsuperscript{\rm 2}School of Information Engineering, Wuhan University of Technology, Wuhan 430070, China\\
    zhongpengzhi@glut.edu.cn,
    mait0917@163.com,
    haolun@glut.edu.cn,
    zhengge0729@glut.edu.cn,\\
    wangjingqi@glut.edu.cn,
    bxy@glut.edu.cn,
    lishuiwang0721@163.com
}

\begin{document}

\maketitle

\begin{abstract}
With sparse and event-driven computation, spiking neural networks show great potential for achieving accurate and power-efficient UAV visual tracking. However, existing SNN-based trackers typically use spike firing rates only for power consumption and lack explicit optimization of actual spike activity. Moreover, regulating spike activity alone does not explicitly encourage stable target representations under partial observations and temporal appearance changes. We propose SBMVTrack, a fully spiking tracking framework that combines spike activity regulation with complementary multi-view representation learning. Specifically, SBMVTrack introduces Energy-Weighted Spike Budgeting (EWSB), which incorporates layer-wise computational costs when regulating spike firing rates and penalizing saturated activations, thereby reducing redundant spike computation. To further improve target representations under the spike budget constraint, we introduce Masked Multi-View Target Modeling (MVTM), which treats the initial template, online template, and search region as temporal views of the same target. By aligning target embeddings between masked and corresponding unmasked views and enforcing cross-view identity consistency, MVTM encourages robustness to missing local cues and temporal appearance changes. Experiments on four UAV benchmarks demonstrate competitive tracking performance with a 24.1\% reduction in estimated power consumption relative to the baseline. On VisDrone2018, SBMVTrack achieves a success rate of 70.0\%, exceeding SpikeTrack by 9.7 percentage points while reducing estimated power consumption by 45.7\%. The source code will be released upon acceptance.
\end{abstract}

\begin{figure}
	\centering
    \includegraphics[width=0.98\columnwidth]{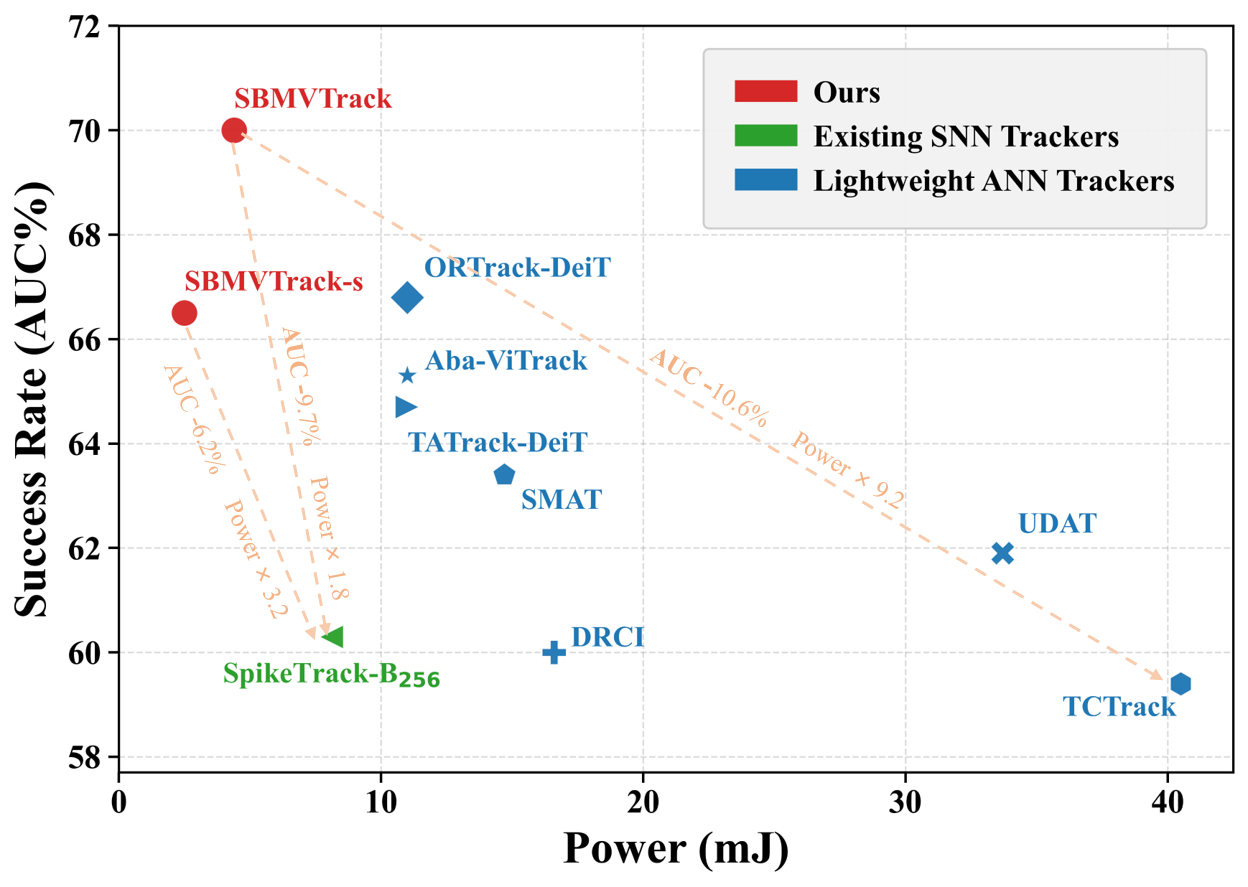}
    \caption{Power–accuracy trade-off on VisDrone2018. SBMVTrack attains 70.0\% AUC at 4.4 mJ inference energy, showing a superior balance between accuracy and efficiency.}
	\label{fig:power}
\end{figure}
\section{Introduction}
Unmanned aerial vehicle (UAV) visual tracking has been widely applied in intelligent surveillance, emergency rescue, autonomous navigation, and low-altitude inspection. However, rapid motion, viewpoint variation, and background interference continue to pose substantial challenges. Meanwhile, limited onboard computational resources and battery capacity require trackers to achieve a favorable balance among accuracy, real-time performance, and power efficiency. In recent years, trackers based on convolutional neural networks and Vision Transformers have significantly improved tracking performance \cite{siamfc++,UDAT,AbaViTrack} . Nevertheless, their reliance on dense multiply-accumulate operations still limits long-term deployment on resource-constrained UAV platforms.

Spiking neural networks (SNNs) \cite{SNN} transmit information through sparse spikes and perform computation in an event-driven manner, offering a promising alternative for building low-power UAV trackers. Existing SNN-based tracking studies mainly follow two directions: RGB image tracking \cite{Spiketrack} and Event-based tracking \cite{SNNTrack,SpikeFET}. Event-based methods leverage the high temporal resolution of event streams to enhance dynamic target modeling, yet their deployment is constrained by the cost of dedicated event cameras. Early RGB-based methods converted Siamese tracking frameworks into spiking form, but did not fully exploit the temporal dynamics of SNNs \cite{Spikingsiamfc++}. More recently, SpikeTrack \cite{Spiketrack} achieved efficient RGB-based spiking tracking through asymmetric time-step design, unidirectional information propagation, and a memory retrieval mechanism. Although these studies have demonstrated the power-efficiency potential of SNNs for visual tracking, their main focus remains on network architecture and feature interaction design. Spike firing rates are typically used only for post-training power estimation or theoretical analysis, rather than being explicitly incorporated into tracking optimization. This raises a key question: can spike activity be explicitly regulated to reduce power consumption without introducing additional inference overhead, while still maintaining competitive tracking performance?

To address this challenge, we propose SBMVTrack, a fully spiking framework for power-efficient UAV tracking. Our framework comprises two key components: \textbf{(i) Energy-Weighted Spike Budgeting (EWSB).}
We introduce EWSB to explicitly regulate spike activity during
training. Since spike activity in different layers incurs different
computational costs, EWSB weights layer-wise activity accordingly
and jointly constrains the energy-weighted firing rate and
saturation activity. This design discourages excessive spike
computation, particularly in computationally expensive layers,
while regulating overall activity around a predefined budget. \textbf{(ii) Masked Multi-View Target Modeling (MVTM).}
Spike budgeting regulates computational activity but does not
explicitly encourage representation stability under missing visual
evidence. Inspired by masked representation learning in
MAE~\citep{MAE} and occlusion-robust tracking in
ORTrack~\citep{ORTrackDeiT}, we introduce MVTM to complement
spike regulation with target representation learning.
MVTM treats the initial template, online template, and search
region as temporal views of the same target and applies masking
to construct partial observations. Masked-to-clean target
embedding alignment encourages stability when local appearance
cues are unavailable, while cross-view identity consistency
encourages stable target identity across temporal appearance
changes. Both EWSB and MVTM are used only during training,
leaving the inference pipeline unchanged and introducing no
additional inference overhead.

Extensive experiments on four UAV benchmarks demonstrate a
favorable power--accuracy trade-off, with a 24.1\% reduction
in estimated power consumption relative to the baseline.
As shown in Fig.~\ref{fig:power}, SBMVTrack achieves the highest
success rate of 70.0\% among the compared trackers on
VisDrone2018 at an estimated power consumption of 4.4~mJ.
Compared with SpikeTrack, it improves the success rate by
9.7 percentage points while reducing estimated power
consumption by 45.7\%.

\section{Related work}
\subsection{Efficient UAV Visual Tracking}
Early UAV visual tracking methods were primarily based on discriminative correlation filters, which enabled efficient target localization through frequency-domain computation \cite{KCF,fDSST}. However, their reliance on handcrafted features limited their representation capability, often leading to performance degradation under complex backgrounds and substantial appearance variations. Subsequently, CNN-based Siamese trackers significantly improved tracking accuracy and robustness by combining deep feature extraction with efficient template matching \cite{siamfc++,SiamRPN++,DRCI}. More recently, Vision Transformers have further advanced UAV tracking by modeling global dependencies between the template and search region through self-attention. To facilitate deployment on resource-constrained platforms, existing studies have introduced lightweight backbones \cite{AbaViTrack}, dynamic token computation \cite{AVTrack}, knowledge distillation \cite{ORTrackDeiT}, and asymmetric feature \cite{AsymTrack} interaction to reduce model complexity and inference cost. Although these methods improve the balance between tracking accuracy and computational efficiency, their inference processes still largely rely on dense multiply–accumulate operations. In this work, we aim to explore spike-driven UAV RGB tracking to achieve power-efficient visual processing.
\begin{figure*}[t]
\centering
\includegraphics[width=1.0\textwidth]{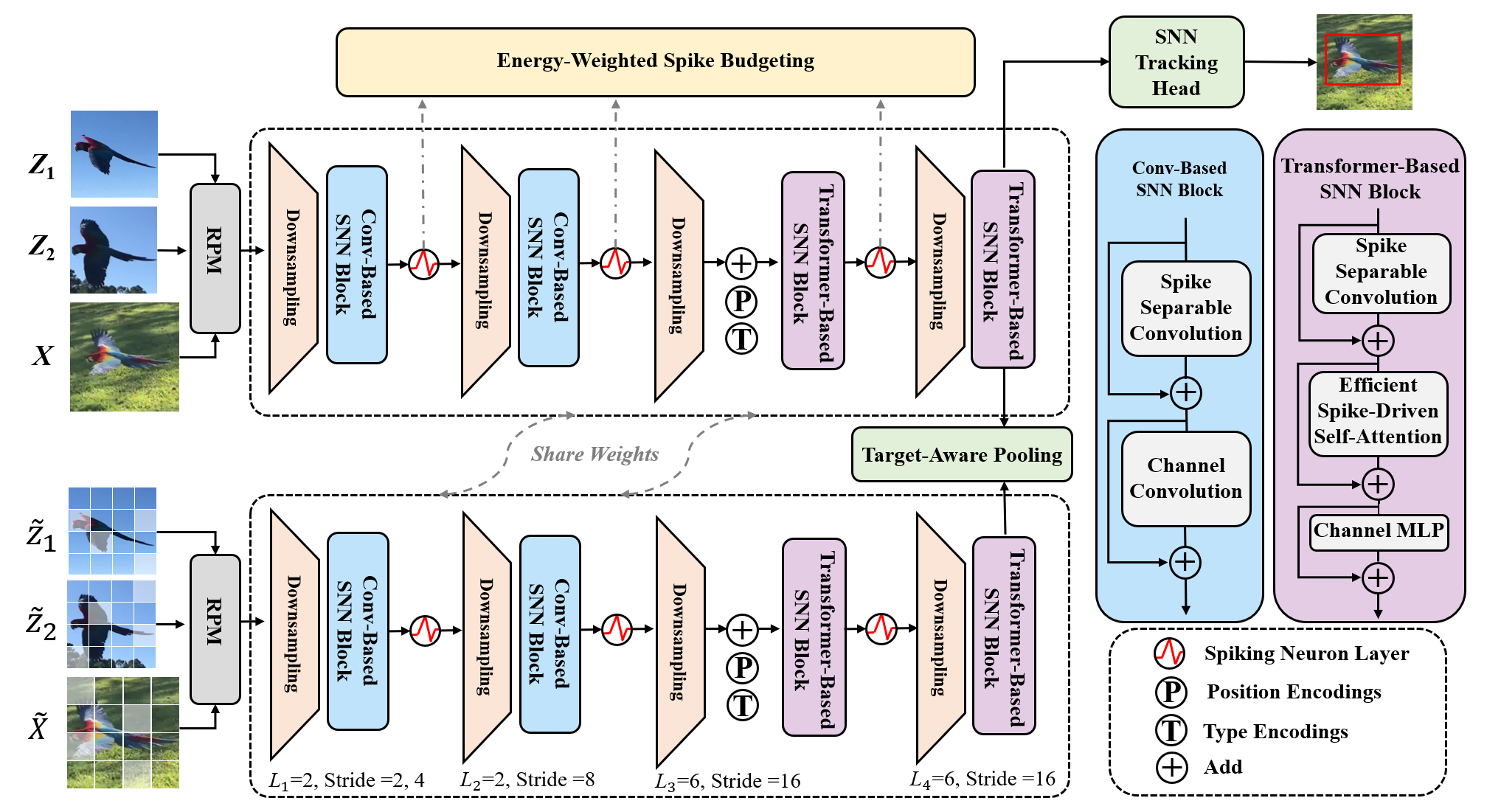}
\caption{Overview of the SBMVTrack architecture. The network comprises an original branch and a masked branch, both processed by a weight-shared spiking backbone. EWSB regulates energy-weighted spike activity, while MVTM performs masked feature reconstruction and cross-view identity-consistency learning through Target-Aware Pooling. During inference, only the original branch, spiking backbone, and SNN tracking head are retained.}
\label{fig:SBMVTrack}
\end{figure*}
\subsection{Spiking Neural Networks for Visual Tracking}
SNN-based visual tracking methods mainly focus on event-camera inputs \cite{SpikeFET,SNNTrack}, leveraging the sparsity and high temporal resolution of event data to achieve promising performance. However, their reliance on dedicated hardware limits practical deployment. For more accessible RGB scenarios, early approaches \cite{Siamsnn,Spikingsiamfc++} still relied on specific ANN frameworks for spike-based conversion, limiting the advantages of fully spiking computation. Recently, SpikeTrack \cite{Spiketrack} further improved the accuracy and
efficiency of RGB spiking tracking through asymmetric time steps and
unidirectional information flow. For UAV tracking scenarios, STATrack
\cite{statrack} enhanced target representation learning by introducing
mutual information maximization. However, these methods mainly focus on improving representation learning or network architecture, while spike firing rates are still used only for post-hoc power estimation rather than explicitly optimized during
training. To address this, we introduce an energy-weighted spike budgeting mechanism from the perspective of spike activity regulation. By constraining the energy-weighted firing rate and saturation activity, it reduces redundant spike computation and lowers power consumption.

\subsection{Multi-View Target Representation Learning}
Multi-view representation learning aims to exploit the consistency and complementarity among different observations of the same target to learn robust feature representations \cite{VGGT}. Recently, masked modeling has been successfully extended to multi-view tasks \cite{Croco, MuM}. In visual tracking, the initial template, online template, and search region naturally constitute multi-view observations of the target along the temporal dimension. However, existing tracking methods typically treat these observations as independent matching inputs and overlook their potential cross-view consistency. Inspired by multi-view learning, we regard different observations during tracking as temporal views of the target and propose Masked Multi-View Target Modeling to explicitly learn cross-view consistent target representations, improving tracking robustness in challenging scenarios.
\section{Method}
\label{sec:method}
This section details the proposed SBMVTrack. We first describe the spiking neuron model, followed by the network architecture. We then introduce EWSB and MVTM. Finally, we present the training objective and inference procedure.

\subsection{Spiking Neuron Model}
\label{sec:spiking_neuron}

We adopt the normalized integer leaky integrate-and-fire (NI-LIF) neuron \cite{Spike2former} to construct the fully spiking tracking network. NI-LIF employs normalized integer-valued activations during training and represents them as equivalent spike sequences during inference, thereby facilitating gradient-based optimization while preserving spike-driven computation. Its neuronal dynamics are formulated as:
\begin{equation}
U[t] = H[t-1] + X[t],
\label{eq:umem}
\end{equation}
\begin{equation}
S[t] = \mathrm{Clip}\!\left(\mathrm{round}\!\left(U[t]\right),\, 0,\, D\right) / D,
\label{eq:spike}
\end{equation}
\begin{equation}
H[t] = \beta \left(U[t] - S[t] \times D\right),
\label{eq:hreset}
\end{equation}
where $X[t]$ denotes the spatial input current at timestep $t$, and $U[t]$ is the membrane potential that integrates the temporal information $H[t-1]$ from the previous timestep with the current spatial input $X[t]$. $S[t]$ denotes the normalized integer-valued output within the range $[0,1]$, and $\mathrm{round}(\cdot)$ represents the rounding operation. $\mathrm{Clip}(x,\min,\max)$ restricts $x$ to the interval $[\min,\max]$, while $D$ is a hyperparameter that determines the maximum integer activation value. The leakage factor $\beta$ controls the retention of the residual membrane potential across timesteps.

\subsection{SBMVTrack Architecture}
\label{sec:architecture}

\noindent\textbf{Fully spiking dual-branch backbone.} As illustrated in Fig. ~\ref{fig:SBMVTrack}, SBMVTrack consists of an original tracking branch and a masked auxiliary branch during training, both of which share the same spiking backbone (E-SpikeFormer)  \cite{ESpikeFormer} and model parameters. The original tracking branch takes an initial template ($Z_1$), an online template ($Z_2$), and a search region ($X$) as inputs, while the masked branch receives the corresponding masked views ($\widetilde{Z}_1$), ($\widetilde{Z}_2$), and ($\widetilde{X}$). Each input is first processed by the Randomized Patch Module (RPM) \cite{SpikeFET} to alleviate padding effects introduced during convolutional feature extraction, and is then fed into the shared spiking backbone.

The backbone is composed of Convolution-based SNN blocks and Transformer-based SNN blocks. After RPM processing, the inputs are encoded by the convolutional SNN module, augmented with positional and type embeddings, and subsequently passed to the Transformer-based SNN module. During training, EWSB collects spike activity from different stages of the backbone and imposes an energy-aware budget constraint to regulate redundant spike computation. Finally, the search features from the original tracking branch are fed into a spiking prediction head to estimate the target center, local offset, and bounding-box size. The masked branch is used only during training to construct auxiliary target representations for MVTM.

\noindent\textbf{Conv-based SNN block.}
Given the input feature \(U\), the convolution-based SNN block performs spatial token mixing and channel-wise interaction through residual spike-driven convolutions:
\begin{align}
U' &= U + SSConv(U), \\
U'' &= U' + ChConv(U').
\end{align}
Here, \(SSConv(\cdot)\) denotes the spike-based separable convolution module for local spatial modeling, while \(ChConv(\cdot)\) performs Channel Convolution. They are formulated as:
\begin{align}
SSConv(U)
&= Conv_{pw}
(\mathcal{SN}(
Conv_{dw}( \mathcal{SN}(\notag \\
&
Conv_{pw}(\mathcal{SN}(U))
)
)
)
), \\
ChConv(U')
&= Conv
(
\mathcal{SN}(
Conv(\mathcal{SN}(U'))
)
).
\end{align}
where \(\mathcal{SN}(\cdot)\) denotes the spiking neuron layer, \(Conv_{pw}(\cdot)\) and \(Conv_{dw}(\cdot)\) denote pointwise and depthwise convolutions, respectively.

\noindent\textbf{Transformer-based SNN block.}
Each Transformer-based SNN block consists of a spike separable convolution, a spike-driven self-attention module (SDSA), and a Channel MLP:
\begin{align}
    U' &= U + SSConv(U), \\
    U'' &= U' + SDSA(U'), \\
    U''' &= U'' + ChMLP(U''), \\
    ChMLP(U'')
    &= Linear
    \left(
    \mathcal{SN}
    \left(
    Linear
    \left(
    \mathcal{SN}(U'')
    \right)
    \right)
    \right).
\end{align}
The SDSA module models global interactions among the initial template, online template, and search region in the spiking feature space, while the channel MLP performs channel-wise feature transformation.

\noindent\textbf{SNN tracking head.}
We employ a lightweight spiking center-based head to estimate the target bounding box \cite{OSTtrack}. After modeling by the spiking backbone, the features corresponding to the search region are selected and fed into three parallel prediction branches. Each branch consists of several Conv-BN-NILIF layers, while the output layer excludes BN and NI-LIF. The three branches predict the target center score   map, local offset, and normalized bounding-box width and height, respectively.

\subsection{Energy-Weighted Spike Budgeting (EWSB)}
\label{sec:ewsb}

Existing SNN-based trackers typically use spike firing rates only for power estimation, without explicitly regulating actual spike activity during training. Moreover, spiking layers incur different computational costs, and their spike activities therefore contribute unequally to the overall power consumption. To address this issue, we propose Energy-Weighted Spike Budgeting (EWSB), which weights layer-wise spike activity according to computational cost and constrains both the energy-weighted firing rate and saturated activity to reduce redundant spike computation. Specifically, let
$\mathbf{S}_l \in \mathbb{R}^{B \times T \times C_l \times H_l \times W_l}$
denote the spike tensor of the $l$-th spiking layer, where $B$, $T$,
$C_l$, $H_l$, and $W_l$ denote the batch size, timesteps, channels, height, and width, respectively. Its average firing rate $r_l$, normalized energy weight $\omega_l$, and the resulting global energy-weighted firing rate $r_{\mathrm{ew}}$ are defined as
\begin{equation}
\begin{aligned}
r_l &=
\frac{1}{B T C_l H_l W_l}
\sum_{b,t,c,h,w}
\mathbf{S}_l^{b,t,c,h,w}, \\
\omega_l &=
\frac{C_l^{\mathrm{comp}}}
{\sum_{j=1}^{L} C_j^{\mathrm{comp}}},
\qquad
r_{\mathrm{ew}} =
\sum_{l=1}^{L} \omega_l r_l ,
\end{aligned}
\label{eq:ew_firing_rate}
\end{equation}
where $C_l^{\mathrm{comp}}$ denotes the computational cost of the
$l$-th layer, and $L$ is the number of spiking layers involved in
EWSB. By assigning larger weights to computationally expensive
layers, $r_{\mathrm{ew}}$ provides a cost-aware measurement of
layer-wise spike activity. We introduce a target spike budget $r_{\mathrm{tar}}$ and employ a
squared penalty to drive the energy-weighted firing rate toward this
predefined budget:
\begin{equation}
\mathcal{L}_{\mathrm{budget}}
=
\left(r_{\mathrm{ew}}-r_{\mathrm{tar}}\right)^2 .
\label{eq:budget_loss}
\end{equation}
Rather than monotonically minimizing spike activity, $\mathcal{L}_{\mathrm{budget}}$ regulates the global energy-weighted firing rate around a predefined budget, thereby suppressing excessive activity while preventing over-sparsification that may impair target representation. Nevertheless,
some redundant responses may still exhibit saturated firing.
We introduce a saturation activity penalty:
\begin{equation}
\mathcal{L}_{\mathrm{sat}}
=
\sum_{l=1}^{L}
\frac{\omega_l}{N_l}
\sum_{i=1}^{N_l}
\sigma\left[
\alpha\left(S_{l,i}-\tau_{\mathrm{sat}}\right)
\right],
\label{eq:sat_loss}
\end{equation}
where $N_l$ is the total number of spike elements in the $l$-th
layer, $\sigma(\cdot)$ denotes the sigmoid function,
$\tau_{\mathrm{sat}}$ is the saturation threshold, and $\alpha$
controls the sharpness of the penalty.

Finally, the EWSB loss is formulated as
\begin{equation}
\mathcal{L}_{\mathrm{EWSB}}
=
\lambda_{\mathrm{budget}}\mathcal{L}_{\mathrm{budget}}
+
\lambda_{\mathrm{sat}}\mathcal{L}_{\mathrm{sat}},
\label{eq:ewsb_loss}
\end{equation}
where $\lambda_{\mathrm{budget}}$ and $\lambda_{\mathrm{sat}}$
control the strengths of the budget constraint and saturation
penalty, respectively.

\subsection{Masked Multi-View Target Modeling}
\label{sec:mvtm}

The initial template ($Z_1$), online template ($Z_2$), and search region ($X$) describe the target's initial identity, recent state, and current observation, respectively, and can therefore be regarded as three correlated temporal views of the same target. To improve robustness against local information loss and appearance variations, we propose Masked Multi-View Target Modeling (MVTM). Specifically, masks sampled from a Cox process \cite{ORTrackDeiT} are applied to $Z_1$, $Z_2$, and $X$ to generate the masked views $\widetilde{Z}_1$, $\widetilde{Z}_2$, and $\widetilde{X}$. The original and masked views are then processed by a weight-shared spiking backbone. To reduce interference from background regions during multi-view learning, we further introduce Target-Aware Pooling (TAP). TAP generates a Gaussian soft weight map according to the target bounding box and performs weighted aggregation over the spatial features around the target:
\begin{equation}
\mathbf{h}_i =
\frac{\sum_{p=1}^{N_i} A_{i,p} \, \mathbf{F}_{i,p}}
{\sum_{p=1}^{N_i} A_{i,p}},
\label{eq:tap}
\end{equation}
where $\mathbf{F}_i$ denotes the feature representation of the $i$-th view, $A_{i,p}$ is the target-aware weight at spatial position $p$, and $N_i$ is the number of spatial locations. The pooled features are then passed through a lightweight projection head composed of Linear-ReLU-Linear, obtaining the original-view embedding $\mathbf{q}_i$ and masked-view embedding $\widetilde{\mathbf{q}}_i$, where $i \in \{1, 2, x\}$. MVTM employs a masked feature reconstruction loss to align each masked view with its corresponding complete view:
\begin{equation}
\mathcal{L}_{\mathrm{rec}} =
\frac{1}{3}
\sum_{i \in \{1,2,x\}}
\left\| \widehat{\widetilde{\mathbf{q}}}_i - \widehat{\mathbf{q}}_i \right\|_2^2,
\label{eq:rec_loss}
\end{equation}
where the $\widehat{.}$ denotes $L_2$ normalization. Meanwhile, the initial template $Z_1$ is treated as the identity anchor, and the online template and search region are constrained to preserve identity consistency with it:
\begin{equation}
\mathcal{L}_{\mathrm{con}} =
\frac{1}{2}
\left(
\left\| \widehat{\mathbf{q}}_2 - \widehat{\mathbf{q}}_1 \right\|_2^2 +
\left\| \widehat{\mathbf{q}}_x - \widehat{\mathbf{q}}_1 \right\|_2^2
\right).
\label{eq:con_loss}
\end{equation}
The overall MVTM objective is defined as
\begin{equation}
\mathcal{L}_{\mathrm{MVTM}} =
\mathcal{L}_{\mathrm{rec}} + \lambda_{\mathrm{con}} \mathcal{L}_{\mathrm{con}},
\label{eq:mvtm_loss}
\end{equation}
where $\lambda_{\mathrm{con}}$ balances masked feature reconstruction and cross-view identity consistency.

\subsection{Training Objective}
\label{sec:training_inference}

\noindent\textbf{Training.}
The overall tracking loss is a weighted sum of a focal classification loss \cite{Focalloss}, an $\ell_1$ regression loss and a generalized IoU loss \cite{Giou}:
\begin{equation}
\mathcal{L}_{\mathrm{track}}
=
\mathcal{L}_{\mathrm{cls}}
+
\lambda_{\mathrm{GIoU}}
\mathcal{L}_{\mathrm{GIoU}}
+
\lambda_{\mathrm{L1}}
\mathcal{L}_{\mathrm{L1}},
\label{eq:track_loss}
\end{equation}
where $\lambda_{\mathrm{GIoU}} = 2$ and $\lambda_{\mathrm{L1}} = 5$. 
The overall training objective is formulated as :
\begin{equation}
\mathcal{L}_{\mathrm{total}}
=
\mathcal{L}_{\mathrm{track}}
+
\lambda_{\mathrm{EWSB}}
\mathcal{L}_{\mathrm{EWSB}}
+
\lambda_{\mathrm{MVTM}}
\mathcal{L}_{\mathrm{MVTM}}.
\label{eq:total_loss}
\end{equation}
where $\lambda_{\mathrm{EWSB}} = 1$  and $\lambda_{\mathrm{MVTM}} = 1$ control the contributions of the energy-weighted spike budgeting and masked multi-view target modeling, respectively.

\begin{table*}[t]
\centering
\setlength{\tabcolsep}{3.6pt}
\begin{tabular}{@{}cc cc cc cc cc cc cc@{}}
\toprule
\multirow{2}{*}{\textbf{Tracker}} & \multirow{2}{*}{\textbf{Source}} & \multicolumn{2}{c}{\textbf{UAVTrack112}} & \multicolumn{2}{c}{\textbf{UAVDT}} & \multicolumn{2}{c}{\textbf{VisDrone2018}} & \multicolumn{2}{c}{\textbf{UAV123}} & \textbf{Power} & \textbf{Param.} \\
\cmidrule(lr){3-4} \cmidrule(lr){5-6} \cmidrule(lr){7-8} \cmidrule(lr){9-10} \cmidrule(lr){11-12}
& & Prec. & Succ. & Prec. & Succ. & Prec. & Succ. & Prec. & Succ.  & \textbf{(mJ)} & \textbf{(M)} \\
\midrule
fDSST \cite{fDSST} & TPAMI 17                       & 56.8 & 39.1 & 66.6 & 38.3 & 69.8 & 51.0 & 58.3 & 40.5 & - & - \\
% ECO\_HC \cite{ECO} & CVPR 17                        & 68.6 & 47.4 & 69.4 & 41.6 & 80.8 & 58.1 & 71.0 & 49.6  & - & - \\
MCCT\_H \cite{MCCT} & CVPR 18                       & 63.4 & 43.6 & 66.8 & 40.2 & 80.3 & 56.7 & 65.9 & 45.7  & - & - \\
ARCF \cite{ARCF} & ICCV 19                          & 67.3 & 45.6 & 72.0 & 45.8 & 79.7 & 58.4 & 67.1 & 46.8  & - & - \\
AutoTrack \cite{AutoTrack} & CVPR 20                & 69.4 & 46.5 & 71.8 & 45.0 & 78.8 & 57.3 & 68.9 & 47.2  & - & - \\
\midrule
HiFT \cite{HiFT} & ICCV 21                          & 74.2 & 57.0 & 65.2 & 47.5 & 71.9 & 52.6 & 78.7 & 59.0 & 33.1 & 9.9 \\
TCTrack \cite{TCTrack} & CVPR 22                    & 76.6 & 59.4 & 72.5 & 53.0 & 79.9 & 59.4 & 80.0 & 60.5 &  40.5 & 9.7 \\
  
AsymTrack \cite{AsymTrack} & AAAI 25                & 81.3 & 66.8 & \underline{84.0} & 59.0 & 83.4 & 60.0 & 76.7 & 59.7  & 8.3 & 3.4 \\

\midrule
Aba-ViTrack \cite{AbaViTrack} & ICCV 23             & 82.6 & 67.6 & 83.4 & 59.9 & 86.1 & 65.3 & 86.4 & 66.4  & 11.0 & 8.0 \\
SMAT \cite{SMAT} & WACV 24                          & 82.2 & 65.3 & 80.8 & 58.7 & 82.5 & 63.4 & 81.8 & 64.6 &  14.7 & 8.6 \\
TATrack \cite{TATrack} & TGRS 24                    & 81.5 & 65.8 & 83.4 & 60.6 & 85.2 & 64.7 & 82.7 & 65.4  & 11.0 & 5.6 \\
AVTrack \cite{AVTrack} & ICML 24                    & 80.3 & 65.4 & 82.1 & 58.7 & 86.0 & 65.3 & 84.8 & 66.8  & 4.5-8.8 & 3.5-7.9 \\
ORTrack \cite{ORTrackDeiT} & CVPR 25                & 83.2 & 67.0 & 83.4 & 60.1 &  \underline{88.6} & \underline{66.8} & 84.3 & 66.4  & 11.0 & 5.8 \\
SGLATrack \cite{SGLATrackDeiT} & CVPR 25            & 82.8 &  67.5 & 81.9 & 59.9 & 80.0 & 61.3 & 84.9 & 66.9 & 7.7 & 7.9 \\
UETrack \cite{UETrack} & CVPR 26                    & 83.0 & \underline{69.2} & 79.6 & 60.4 & 80.7 & 62.2 & 86.6 & \underline{67.1}  & 8.5 & 6.4 \\
\midrule
SpikeTrack \cite{Spiketrack} & CVPR 26              & \textbf{85.3} & \underline{69.2} & 75.9 & 56.3 & 80.2 & 60.3 & \textbf{88.7} & 67.5  & 8.1 & 36.8 \\

\midrule
\textbf{SBMVTrack} & \multirow{2}{*}{\textbf{Ours}} & 82.8  & \textbf{70.1}  & \textbf{84.3} & \textbf{64.8}  &\textbf{90.0} & \textbf{70.0} &\underline{87.6} & \textbf{68.0} &  \underline{4.4}  & 12.6\\
\textbf{SBMVTrack-S} &  & \underline{83.8}  & \textbf{70.1}  & 83.1 & \underline{63.8} & 85.3 & 66.5 & 85.6 & 66.6 & \textbf{2.5}  & 7.4 \\
\bottomrule
\end{tabular}
\caption{Comparison of SBMVTrack with representative trackers in terms of Precision (Prec.), Success rate (Succ.), Power (mJ), and Parameters (Params.) on UAVTrack112, UAVDT, VisDrone2018, and UAV123. The best and second-best results for each metric are highlighted in bold and underlined, respectively. The compared trackers are categorized into DCF-based, CNN-based, ViT-based, and SNN-based methods.}
\label{tab:sota}
\end{table*}

\section{Experiments}
\subsection{Implementation Details}

\noindent\textbf{Model.}
We construct two SBMVTrack variants with different backbone scales to evaluate the trade-off between tracking performance and power efficiency, namely SBMVTrack and SBMVTrack-S, which employ E-SpikeFormer-10M and E-SpikeFormer-5.1M backbones \cite{ESpikeFormer}, respectively. The template and search resolutions are set to $128\times128$ and $256\times256$, respectively. All NI-LIF neurons operate with a single timestep ($T=1$), and the spike quantization level is set to $D=4$. All experiments run on a PC with an i9-10850K processor, 16GB RAM, and an NVIDIA TitanX GPU.

\noindent\textbf{Training strategy.}
We train SBMVTrack on a combination of LaSOT~\cite{lasot}, GOT-10k~\cite{got10k}, COCO~\cite{coco}, and TrackingNet~\cite{trackingnet}. The batch size is set to 16, and AdamW \cite{Adamw} is adopted with a weight decay of $1\times10^{-4}$. The initial learning rates of the prediction head and backbone are set to $6\times10^{-4}$ and $6\times10^{-5}$, respectively. The model is trained for 300 epochs with 60,000 sample pairs per epoch, using a linear warm-up for the first 5 epochs and reducing the learning rate to 0.1× at the 240th epoch.

\noindent\textbf{Hyperparameter settings.}
For EWSB, the target energy-weighted firing rate is set to $r_{\mathrm{tar}}=0.12$, and the spike-budget loss weight is set to $\lambda_{\mathrm{bud}}=1.0$. The saturation threshold and saturation penalty weight are set to $\tau_{\mathrm{sat}}=0.9$ and $\lambda_{\mathrm{sat}}=0.5$, respectively. The sharpness coefficient is set to $\alpha = 10.0$. For MVTM, the weight of the cross-view identity-consistency loss is set to $\lambda_{\mathrm{con}}=0.5$. Unless otherwise specified, these settings are used in all experiments.

\noindent\textbf{Inference settings.}
During inference, only the original tracking branch, spiking backbone, and SNN tracking head are retained, while the masked branch, TAP, and projection head used during training are removed. The initial template remains fixed throughout tracking. The online template is updated every 25 frames, but only when the prediction confidence exceeds 0.7 is it replaced by the current target region. Following prior SNN studies~\cite{Spiketrack, ESpikeFormer}, we report the average power consumption (mJ) on UAVTrack112~\cite{uavtrack112}, UAVDT~\cite{UAVDT}, VisDrone2018~\cite{Visdrone2018}, and UAV123~\cite{UAV123}, calculation details are provided in the supplementary material.

\begin{table}[t]
\centering
\setlength{\tabcolsep}{4.2pt}
\begin{tabular}{ccccccc}
\toprule
\multirow{2}{*}{EWSB}
& \multirow{2}{*}{MVTM}
& \multicolumn{2}{c}{UAV123}
& \multicolumn{2}{c}{VisDrone2018}
& Power\\
\cmidrule(lr){3-4}
\cmidrule(lr){5-6}

& & Prec. & Succ. & Prec. & Succ. & (mJ)\\
\midrule
              &       & 86.1          & 67.0          & 82.5          & 65.0          & 5.8 \\
\checkmark    &   & 87.1          & 67.9          & 84.2 &   65.8        & \textbf{4.4} \\
              & \checkmark   & 87.4    & \textbf{68.0}          & 84.8          & 66.7          & 5.8 \\
\checkmark    & \checkmark   & \textbf{87.6} & \textbf{68.0} & \textbf{90.0}            & \textbf{70.0}            & \textbf{4.4} \\
\bottomrule
\end{tabular}
\caption{Ablation study of EWSB and MVTM on UAV123 and VisDrone2018.}
\label{tab:component_ablation}
\end{table}

\subsection{Comparison with State-of-the-Art Trackers}
Tab.~\ref{tab:sota} compares SBMVTrack with 15 representative trackers on four UAV benchmarks, including traditional DCF-based, CNN-based, and recent ViT-based methods, as well as the SNN-based tracker SpikeTrack. SBMVTrack consistently ranks among the top two in precision (Prec.), success rate (Succ.), and power efficiency (Power). Specifically, SBMVTrack attains a Succ. of 70.1\% on UAVTrack112, outperforming all compared methods, and achieves 84.3\% Prec. and 64.8\% Succ. on UAVDT. On VisDrone2018, it achieves the highest Prec. and Succ. of 90.0\% and 70.0\%, respectively. On UAV123, it obtains a competitive Prec. of 87.6\% and the highest Succ. of 68.0\%. Compared with the recent ANN-based lightweight trackers such as UETrack, AVTrack, and ORTrack, SBMVTrack achieves superior or comparable accuracy with significantly lower power consumption. For instance, SBMVTrack consumes only 4.4 mJ, which is 48.2\% less than UETrack (8.5 mJ) and 60.0\% less than ORTrack (11.0 mJ), while outperforming them on most benchmarks. Compared with SpikeTrack, SBMVTrack reduces estimated power consumption from 8.1~mJ to 4.4~mJ, a reduction of 45.7\%, while improving the success rate on VisDrone2018 from 60.3\% to 70.0\%, a gain of 9.7 percentage points. SBMVTrack-S further reduces the power consumption to 2.5~mJ while maintaining competitive tracking accuracy. These results demonstrate that SBMVTrack achieves a more favorable trade-off between tracking results and power efficiency compared to both ANN-based and existing SNN-based lightweight trackers.

\begin{figure}[t]
    \centering
    \includegraphics[width=\columnwidth]{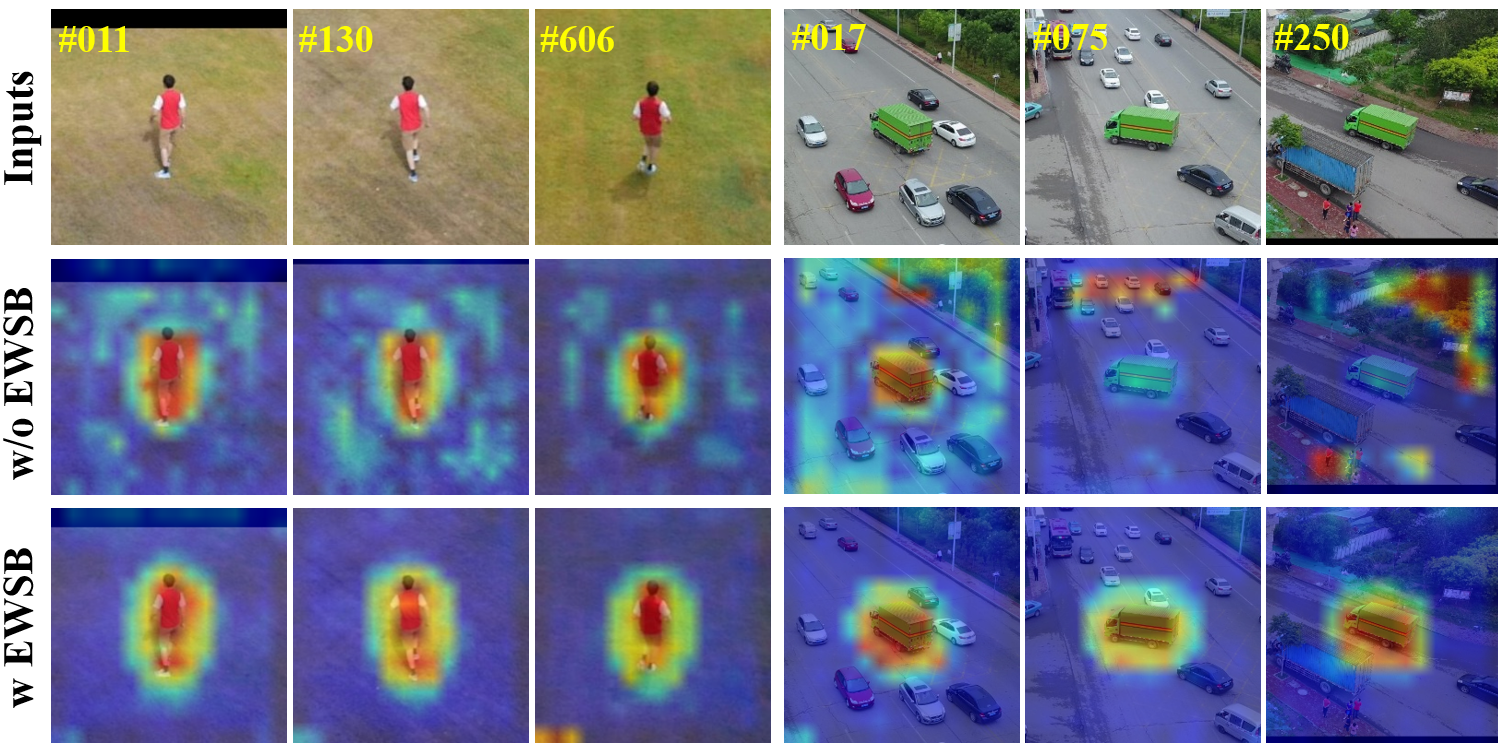}
    \caption{Comparison of search feature activation maps. The second and third rows show the activation maps produced by SBMVTrack without and with EWSB, respectively.}
    \label{fig:activation}
\end{figure}

\subsection{Ablation Study}
\noindent\textbf{Effectiveness of EWSB and MVTM.}
Table~\ref{tab:component_ablation} evaluates the contributions of EWSB and MVTM. EWSB reduces the power consumption from 5.8 mJ to 4.4 mJ while improving tracking performance. MVTM further enhances target representation learning without additional inference cost. Their combination achieves the best overall performance on both benchmarks with a 24.1\% power reduction, demonstrating their complementary effects.

\begin{table}[t]
\centering

\begin{tabular}{lccc}
\toprule
Method & Prec. & Succ.& Power (mJ)\\
\midrule
Baseline        & 86.1 & 67.0 & 5.8 \\
Unweighted Budget   & 85.5 & 66.9 & 4.8 \\
Energy-Weighted Budget & 86.5 & 67.4 & 5.1 \\
EWSB            & \textbf{87.1} & \textbf{67.9} & \textbf{4.4} \\
\bottomrule
\end{tabular}
\caption{Ablation study of the key designs in EWSB on UAV123.}
\label{tab:ewsb_ablation}
\end{table}

\begin{table}[t]
\centering
\setlength{\tabcolsep}{4.5pt}
\begin{tabular}{cccccccc}
\toprule
\multirow{2}{*}{$r_{\mathrm{tar}}$}
& \multicolumn{2}{c}{UAV123}
& \multicolumn{2}{c}{UAVDT}
& \multirow{2}{*}{\shortstack{Avg. \\SFR $\downarrow$}}
& \multirow{2}{*}{\shortstack{Power\\(mJ) $\downarrow$}} \\
\cmidrule(lr){2-3}
\cmidrule(lr){4-5}
& Succ. & Prec.
& Succ. & Prec.
& & \\
\midrule
Baseline & 67.0 & 86.1 & 61.2 & 78.5 & 0.188 & 5.80 \\
0.08     & \textbf{68.8} & \textbf{88.7}
         & 61.9 & 79.9 & \textbf{0.131} & \textbf{4.12} \\
0.10     & 67.4 & 86.4
         & 61.8 & 80.2 & 0.133 & \underline{4.21} \\
0.12     & \underline{67.9} & \underline{87.1}
         & \textbf{64.1} & \textbf{82.9} & 0.140 & 4.42 \\
0.14     & 66.6 & 85.3
         & \underline{63.2} & \underline{80.8} & 0.146 & 4.61 \\
0.16     & 67.0 & 86.0
         & 62.9 & \underline{80.8} & 0.153 & 4.77 \\
\bottomrule
\end{tabular}
\caption{Impact of $r_{\mathrm{tar}}$ on tracking accuracy,
average spike firing rate (SFR), and power consumption.}
\label{tab:rtar_analysis}
\end{table}

\begin{table}[t]
\centering
\begin{tabular}{cccccc}
\toprule
\multirow{2}{*}{Recon.}
& \multirow{2}{*}{Cons.}
& \multicolumn{2}{c}{UAV123}
& \multicolumn{2}{c}{VisDrone2018} \\
\cmidrule(lr){3-4} \cmidrule(lr){5-6}
& &  Prec. &Succ. & Prec.& Succ. \\
\midrule
               &              & 86.1 & 67.0 & 82.5 & 65.0 \\
\checkmark     &              & 84.9 & 66.4 & 80.8 & 63.3 \\
               & \checkmark   & 85.3 & 66.6 & 84.4 & 65.5 \\
\checkmark     & \checkmark   & \textbf{87.4} & \textbf{68.0} & \textbf{84.8} & \textbf{66.7} \\
\bottomrule
\end{tabular}
\caption{Ablation study of the key objectives in MVTM on UAV123 and VisDrone2018. }
\label{tab:mvtm_ablation}
\end{table}

\begin{figure}[!t]
    \centering
    \includegraphics[width=\columnwidth]{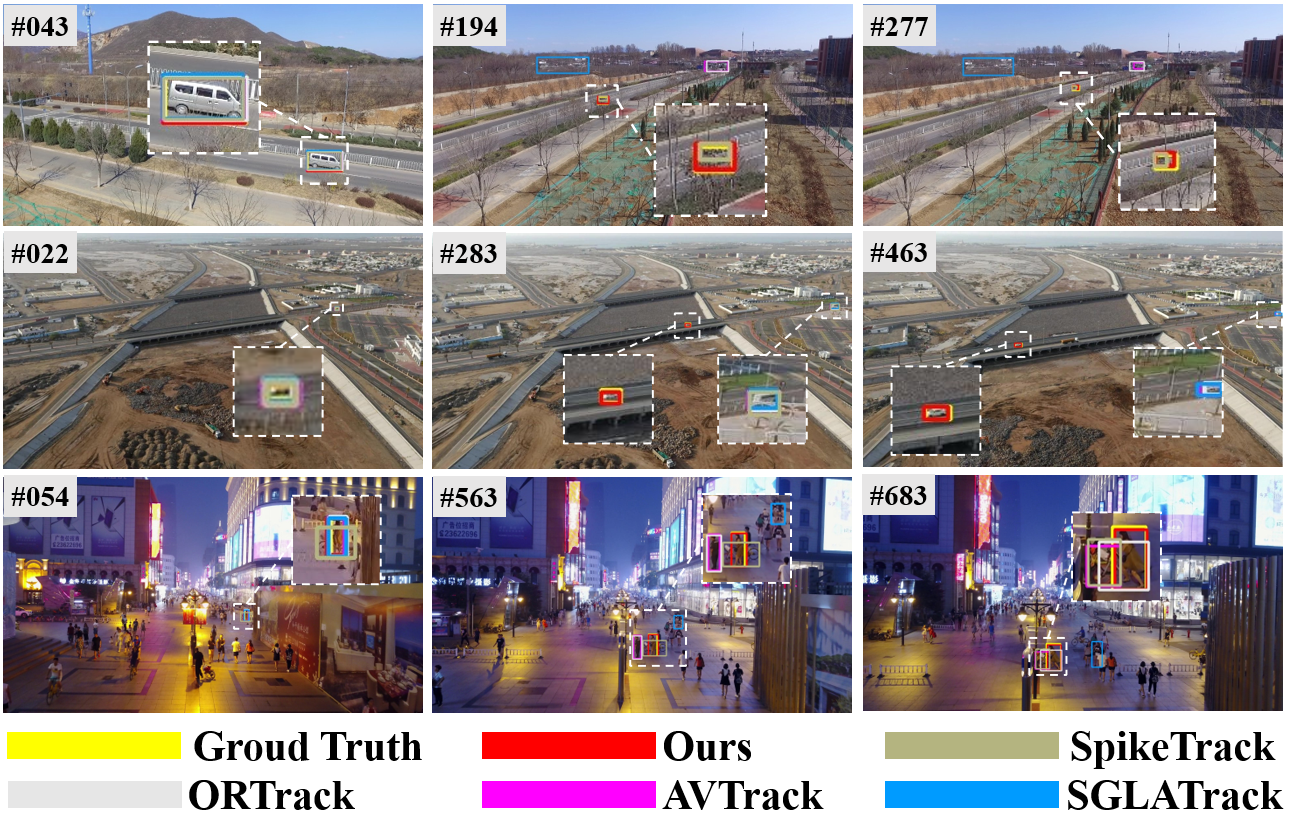}
    \caption{Qualitative evaluation on three video sequences from UAVDT, UAV123, and VisDrone2018 (i.e., S1101, car15, and uav0000074\_04320\_s).}
    \label{fig:track}
\end{figure}
\begin{table}[t]
\centering
\begin{tabular}{ccccccc}
\toprule
\multirow{2}{*}{$\lambda_{\mathrm{con}}$} 
& \multicolumn{2}{c}{UAV123} 
& \multicolumn{2}{c}{VisDrone2018} 
& \multicolumn{2}{c}{UAVDT} \\
\cmidrule(lr){2-3} \cmidrule(lr){4-5} \cmidrule(lr){6-7}
& Prec. & Succ. & Prec. & Succ. & Prec. & Succ. \\
\midrule
0.1 & 87.3 & 67.8 & 83.7 & 65.9 & 77.3 & 60.4 \\
0.5 & \textbf{87.4} & \textbf{68.0} & \textbf{84.8} & \textbf{66.7} & 79.7 & 61.6 \\
1.0 & 86.9 & 67.3 & 84.6 & 66.3 & \textbf{80.9} & \textbf{62.8} \\
1.5 & 86.9 & 67.3 & 82.7 & 65.1 & 78.2 & 60.5 \\
\bottomrule
\end{tabular}
\caption{Impact of different values of $\lambda_{\mathrm{con}}$.}
\label{tab:lambda_con}
\end{table}

\begin{table}[!t]
\centering
\setlength{\tabcolsep}{3.5pt}
\begin{tabular}{lccccc}
\toprule
Method & GPU & CPU & Prec. & Succ.  &Power (mJ) \\
\midrule
OSTrack       & 65.4  & 5.9  & 84.2 & 64.8 & 98.9 \\
ORTrack       & \textbf{226.4} & \textbf{55.4}
              & 88.6 & 66.8 & 11.0 \\
SpikeTrack    & 60.4  & 12.1 & 80.2 & 60.3 & 8.1 \\
\midrule
SBMVTrack     & 118.8 & 24.4
              & \textbf{90.0} & \textbf{70.0} & 4.4 \\
SBMVTrack-S   & 157.0 & 33.3 & 85.3 & 66.5 & \textbf{2.5} \\
\bottomrule
\end{tabular}
\caption{Inference speed, tracking accuracy, and theoretical power consumption on VisDrone2018.}
\label{tab:speed}
\end{table}

\noindent\textbf{Analysis of EWSB.}
Tab.~\ref{tab:ewsb_ablation} evaluates energy weighting and saturation regulation on UAV123. The unweighted budget reduces power consumption but slightly decreases tracking accuracy. The energy-weighted budget improves precision and success rate over the baseline while reducing power consumption from 5.8 to 5.1~mJ. Adding saturation regulation further improves precision and success rate to 87.1\% and 67.9\%, respectively, while lowering estimated power consumption to 4.4~mJ.

\noindent\textbf{Impact of $r_{\mathrm{tar}}$.}
Tab.~\ref{tab:rtar_analysis} analyzes different target firing-rate budgets. A smaller $r_{\mathrm{tar}}$ generally reduces the average spike firing rate and power consumption, whereas an overly restrictive budget may degrade tracking performance, especially on UAVDT. We set $r_{\mathrm{tar}}=0.12$, which reduces Avg. SFR from 0.188 to 0.140 and power consumption from 5.80 mJ to 4.42 mJ, while improving tracking performance on both benchmarks.

\noindent\textbf{Analysis of MVTM.}
Tab.~\ref{tab:mvtm_ablation} evaluates the two learning objectives
in MVTM. Using either reconstruction or consistency alone does not
consistently improve tracking performance. In contrast, combining
the two objectives achieves clear gains on both benchmarks,
indicating that reconstruction and cross-view consistency provide
complementary supervision for robust target representation learning.

\noindent\textbf{Impact of $\lambda_{\mathrm{con}}$.}
Tab.~\ref{tab:lambda_con} studies the influence of the
cross-view consistency weight. $\lambda_{\mathrm{con}}=0.5$
achieves the best overall performance, while a smaller value provides
insufficient regularization and an excessively large value may
over-constrain temporal appearance variations.

\noindent\textbf{Speed Evaluation.}
As shown in Tab.~\ref{tab:speed}, SBMVTrack achieves higher accuracy
and lower power consumption than the compared trackers. The lightweight SBMVTrack-S achieves 157.0 FPS on the GPU and 33.3 FPS on the CPU, demonstrating inference throughput above 30 FPS on the evaluated CPU without requiring specialized neuromorphic hardware.

\noindent\textbf{Qualitative Results.}
As shown in Fig.~\ref{fig:activation}, the baseline produces dispersed responses over background regions, whereas SBMVTrack concentrates its activations more clearly around the target, indicating more discriminative target representations. Fig.~\ref{fig:track} compares SBMVTrack with four advanced UAV trackers on representative sequences from UAVDT, UAV123, and VisDrone2018. Under challenging conditions such as background clutter, similar-object, and small targets, SBMVTrack maintains stable localization, whereas several competing methods fail to track the target.

\noindent\textbf{Limitation.}
The power efficiency of SBMVTrack is currently evaluated theoretically based on operation counts and spike activity, while its practical efficiency may vary with hardware architectures, memory access, and specific implementation details. In future work, we plan to further validate its actual power consumption and runtime efficiency on neuromorphic hardware platforms.

\section{Conclusion}
We presented SBMVTrack, a power-efficient fully spiking framework for UAV tracking. SBMVTrack explicitly regulates layer-wise spike activity through Energy-Weighted Spike Budgeting (EWSB) and improves the stability of target representations through Masked Multi-View Target Modeling (MVTM). Extensive experiments on multiple UAV tracking benchmarks demonstrate that SBMVTrack achieves performance competitive with advanced ANN trackers while substantially reducing theoretical power consumption, resulting in a favorable accuracy-power trade-off. We hope this work provides a new perspective on energy-aware training of SNNs for UAV vision tasks.

\bibliography{main}

@article{KCF,
  title={High-speed tracking with kernelized correlation filters},
  author={Henriques, Jo{\~a}o F and Caseiro, Rui and Martins, Pedro and Batista, Jorge},
  journal={IEEE TPAMI},
  volume={37},
  number={3},
  pages={583--596},
  year={2014},
  publisher={IEEE}
}

@inproceedings{ARCF,
  title={Learning aberrance repressed correlation filters for real-time UAV tracking},
  author={Huang, Ziyuan and Fu, Changhong and Li, Yiming and Lin, Fuling and Lu, Peng},
  booktitle={ICCV},
  pages={2891--2900},
  year={2019}
}

@inproceedings{AutoTrack,
  title={AutoTrack: Towards high-performance visual tracking for UAV with automatic spatio-temporal regularization},
  author={Li, Yiming and Fu, Changhong and Ding, Fangqiang and Huang, Ziyuan and Lu, Geng},
  booktitle={CVPR},
  pages={11923--11932},
  year={2020}
}

@inproceedings{HiFT,
  title={Hift: Hierarchical feature transformer for aerial tracking},
  author={Cao, Ziang and Fu, Changhong and Ye, Junjie and Li, Bowen and Li, Yiming},
  booktitle={ICCV},
  pages={15457--15466},
  year={2021}
}

@inproceedings{UDAT,
  title={Unsupervised domain adaptation for nighttime aerial tracking},
  author={Ye, Junjie and Fu, Changhong and Zheng, Guangze and Paudel, Danda Pani and Chen, Guang},
  booktitle={CVPR},
  pages={8896--8905},
  year={2022}
}

@inproceedings{TCTrack,
  title={TCTrack: Temporal contexts for aerial tracking},
  author={Cao, Ziang and Huang, Ziyuan and Pan, Liang and Zhang, Shiwei and Liu, Ziwei and Fu, Changhong},
  booktitle={CVPR},
  pages={14798--14808},
  year={2022}
}

@inproceedings{AbaViTrack,
  title={Adaptive and background-aware vision transformer for real-time uav tracking},
  author={Li, Shuiwang and Yang, Yangxiang and Zeng, Dan and Wang, Xucheng},
  booktitle={ICCV},
  pages={13989--14000},
  year={2023}
}

@inproceedings{SMAT,
  title={Separable self and mixed attention transformers for efficient object tracking},
  author={Gopal, Goutam Yelluru and Amer, Maria A},
  booktitle={WACV},
  pages={6708--6717},
  year={2024}
}

@inproceedings{ORTrackDeiT,
  title={Learning Occlusion-Robust Vision Transformers for Real-Time UAV Tracking},
  author={Wu, You and Wang, Xucheng and Yang, Xiangyang and Liu, Mengyuan and Zeng, Dan and Ye, Hengzhou and Li, Shuiwang},
  booktitle={CVPR},
  pages={17103--17113},
  year={2025}
}

@inproceedings{SGLATrackDeiT,
  title={Similarity-guided layer-adaptive vision transformer for UAV tracking},
  author={Xue, Chaocan and Zhong, Bineng and Liang, Qihua and Zheng, Yaozong and Li, Ning and Xue, Yuanliang and Song, Shuxiang},
  booktitle={CVPR},
  pages={6730--6740},
  year={2025}
}

@article{SNNTrack,
  title={Spiking neural networks with adaptive membrane time constant for event-based tracking},
  author={Zhang, Jiqing and Zhang, Malu and Wang, Yuanchen and Liu, Qianhui and Yin, Baocai and Li, Haizhou and Yang, Xin},
  journal={IEEE TIP},
  volume={34},
  number={},
  pages={1009-1021},
  year={2025},
  publisher={IEEE}
}

@article{TATrack,
  title={Learning target-aware vision transformers for real-time UAV tracking},
  author={Li, Shuiwang and Yang, Xiangyang and Wang, Xucheng and Zeng, Dan and Ye, Hengzhou and Zhao, Qijun},
  journal={IEEE TGRS},
  volume={62},
  pages={1--18},
  year={2024},
  publisher={IEEE}
}

@article{ESpikeFormer,
  title={Scaling spike-driven transformer with efficient spike firing approximation training},
  author={Yao, Man and Qiu, Xuerui and Hu, Tianxiang and Hu, Jiakui and Chou, Yuhong and Tian, Keyu and Liao, Jianxing and Leng, Luziwei and Xu, Bo and Li, Guoqi},
  journal={IEEE TPAMI},
  volume={47},
  number={4},
  pages={2973-2990},
  year={2025},
  publisher={IEEE}
}

@inproceedings{SpikeFET,
  title={Fully Spiking Neural Networks for Unified Frame-Event Object Tracking},
  author={Yang, Jingjun and Fan, Liangwei and Zhang, Jinpu and Lian, Xiangkai and Shen, Hui and Hu, Dewen},
  booktitle={NeurIPS},
  year={2025}
}

@article{Spikingsiamfc++,
  title={Spiking siamfc++: Deep spiking neural network for object tracking},
  author={Xiang, Shuiying and Zhang, Tao and Jiang, Shuqing and Han, Yanan and Zhang, Yahui and Guo, Xingxing and Yu, Licun and Shi, Yuechun and Hao, Yue},
  journal={Nonlinear dynamics},
  volume={112},
  number={10},
  pages={8417--8429},
  year={2024},
  publisher={Springer}
}

@inproceedings{AVTrack,
  title={Learning Adaptive and View-Invariant Vision Transformer for Real-Time UAV Tracking},
  author={Li, Yongxin and Liu, Mengyuan and Wu, You and Wang, Xucheng and Yang, Xiangyang and Li, Shuiwang},
  booktitle={ICML},
  year={2024},
  pages={28403--28420}
}

@article{fDSST,
  title={Discriminative scale space tracking},
  author={Danelljan, Martin and H{\"a}ger, Gustav and Khan, Fahad Shahbaz and Felsberg, Michael},
  journal={IEEE TPAMI},
  volume={39},
  number={8},
  pages={1561--1575},
  year={2016},
  publisher={IEEE}
}

@inproceedings{siamfc++,
  title={SiamFC++: Towards robust and accurate visual tracking with target estimation guidelines},
  author={Xu, Yinda and Wang, Zeyu and Li, Zuoxin and Yuan, Ye and Yu, Gang},
  booktitle={AAAI},
  volume={34},
  pages={12549--12556},
  year={2020}
}

@inproceedings{OSTtrack,
  title={Joint feature learning and relation modeling for tracking: A one-stream framework},
  author={Ye, Botao and Chang, Hong and Ma, Bingpeng and Shan, Shiguang and Chen, Xilin},
  booktitle={ECCV},
  pages={341--357},
  year={2022},
}

@inproceedings{UAVDT,
  title={The unmanned aerial vehicle benchmark: Object detection and tracking},
  author={Du, Dawei and Qi, Yuankai and Yu, Hongyang and Yang, Yifan and Duan, Kaiwen and Li, Guorong and Zhang, Weigang and Huang, Qingming and Tian, Qi},
  booktitle={ECCV},
  pages={370--386},
  year={2018}
}

@inproceedings{Visdrone2018,
  title={Visdrone-sot2018: The vision meets drone single-object tracking challenge results},
  author={Wen, Longyin and Zhu, Pengfei and Du, Dawei and Bian, Xiao and Ling, Haibin and Hu, Qinghua and Liu, Chenfeng and Cheng, Hao and Liu, Xiaoyu and Ma, Wenya and others},
  booktitle={ECCV},
  pages={0--0},
  year={2018}
}

@inproceedings{UAV123,
  title={A Benchmark and Simulator for UAV Tracking},
  author={Matthias Mueller and Neil G. Smith and Bernard Ghanem},
  booktitle={ECCV},
  volume={7},
  pages={445--461},
  year={2016}
}

@inproceedings{DRCI,
  title={Towards discriminative representations with contrastive instances for real-time uav tracking},
  author={Zeng, Dan and Zou, Mingliang and Wang, Xucheng and Li, Shuiwang},
  booktitle={ICME},
  pages={1349--1354},
  year={2023},
}

@inproceedings{Focalloss,
  title={Cornernet: Detecting objects as paired keypoints},
  author={Law, Hei and Deng, Jia},
  booktitle={ECCV},
  pages={734--750},
  year={2018}
}

@inproceedings{Giou,
  title={Generalized intersection over union: A metric and a loss for bounding box regression},
  author={Rezatofighi, Hamid and Tsoi, Nathan and Gwak, JunYoung and Sadeghian, Amir and Reid, Ian and Savarese, Silvio},
  booktitle={CVPR},
  pages={658--666},
  year={2019}
}

@article{Lasot,
  title={Lasot: A high-quality large-scale single object tracking benchmark},
  author={Fan, Heng and Bai, Hexin and Lin, Liting and Yang, Fan and Chu, Peng and Deng, Ge and Yu, Sijia and Harshit and Huang, Mingzhen and Liu, Juehuan and others},
  journal={IJCV},
  volume={129},
  pages={439--461},
  year={2021},
  publisher={Springer}
}

@article{got10k,
  title={Got-10k: A large high-diversity benchmark for generic object tracking in the wild},
  author={Huang, Lianghua and Zhao, Xin and Huang, Kaiqi},
  journal={IEEE TPAMI},
  number={5},
  pages={1562--1577},
  year={2019},
  publisher={IEEE}
}

@inproceedings{Trackingnet,
  title={Trackingnet: A large-scale dataset and benchmark for object tracking in the wild},
  author={Muller, Matthias and Bibi, Adel and Giancola, Silvio and Alsubaihi, Salman and Ghanem, Bernard},
  booktitle={ECCV},
  pages={300--317},
  year={2018}
}

@inproceedings{coco,
  title={Microsoft coco: Common objects in context},
  author={Lin, Tsung-Yi and Maire, Michael and Belongie, Serge and Hays, James and Perona, Pietro and Ramanan, Deva and Doll{\'a}r, Piotr and Zitnick, C Lawrence},
  booktitle={ECCV},
  pages={740--755},
  year={2014},
}

@inproceedings{SiamRPN++,
  title={Siamrpn++: Evolution of siamese visual tracking with very deep networks},
  author={Li, Bo and Wu, Wei and Wang, Qiang and Zhang, Fangyi and Xing, Junliang and Yan, Junjie},
  booktitle={CVPR},
  pages={4282--4291},
  year={2019}
}

@inproceedings{MCCT,
	title="Multi-cue Correlation Filters for Robust Visual Tracking",
	author="Ning {Wang} and Wengang {Zhou} and Qi {Tian} and Richang {Hong} and Meng {Wang} and Houqiang {Li}",
	booktitle="CVPR",
	pages="4844--4853",
	year="2018"
}

@inproceedings{Spiketrack,
  title={Spiketrack: A spike-driven framework for efficient visual tracking},
  author={Zhang, Qiuyang and Cheng, Jiujun and Mao, Qichao and Liu, Cong and Fang, Yu and Li, Yuhong and Ge, Mengying and Gao, Shangce},
  booktitle={CVPR},
  pages={6802--6811},
  year={2026}
}

@inproceedings{UETrack,
  title={UETrack: A Unified and Efficient Framework for Single Object Tracking},
  author={Kang, Ben and Zhao, Jie and Chen, Xin and Geng, Wanting and Zhang, Bin and Zhang, Lu and Wang, Dong and Lu, Huchuan},
  booktitle={CVPR},
  pages={20890--20901},
  year={2026}
}

@inproceedings{AsymTrack,
  title={Two-stream Beats One-stream: Asymmetric Siamese Network for Efficient Visual Tracking}, 
  author={Zhu, Jiawen and Tang, Huayi and Chen, Xin and Wang, Xinying and Wang, Dong and Lu, Huchuan},
  booktitle={AAAI},
  pages = {10959–10967},
  year={2025}
}

@article{SNN,
title = {Networks of spiking neurons: The third generation of neural network models},
  author={Wolfgang Maass},
  journal={Neural Networks},
  volume={10},
  pages={1659-1671},
  year={1997}
}

@inproceedings{Siamsnn,
  title={Siamsnn: Siamese spiking neural networks for energy-efficient object tracking},
  author={Luo, Yihao and Xu, Min and Yuan, Caihong and Cao, Xiang and Zhang, Liangqi and Xu, Yan and Wang, Tianjiang and Feng, Qi},
  booktitle={International conference on artificial neural networks},
  pages={182--194},
  year={2021},
  organization={Springer}
}

@article{Croco,
  title={Croco: Self-supervised pre-training for 3d vision tasks by cross-view completion},
  author={Weinzaepfel, Philippe and Leroy, Vincent and Lucas, Thomas and Br{\'e}gier, Romain and Cabon, Yohann and Arora, Vaibhav and Antsfeld, Leonid and Chidlovskii, Boris and Csurka, Gabriela and Revaud, J{\'e}r{\^o}me},
  journal={Advances in Neural Information Processing Systems},
  volume={35},
  pages={3502--3516},
  year={2022}
}

@inproceedings{MuM,
  title={MuM: Multi-View Masked Image Modeling for 3D Vision},
  author={Nordstr{\"o}m, David and Edstedt, Johan and Kahl, Fredrik and B{\"o}kman, Georg},
  booktitle={CVPR},
  pages={21736--21747},
  year={2026}
}

@inproceedings{VGGT,
  title={Vggt: Visual geometry grounded transformer},
  author={Wang, Jianyuan and Chen, Minghao and Karaev, Nikita and Vedaldi, Andrea and Rupprecht, Christian and Novotny, David},
  booktitle={Proceedings of the Computer Vision and Pattern Recognition Conference},
  pages={5294--5306},
  year={2025}
}

@inproceedings{Spike2former,
  title={Spike2former: Efficient spiking transformer for high-performance image segmentation},
  author={Lei, Zhenxin and Yao, Man and Hu, Jiakui and Luo, Xinhao and Lu, Yanye and Xu, Bo and Li, Guoqi},
  booktitle={Proceedings of the AAAI Conference on Artificial Intelligence},
  volume={39},
  number={2},
  pages={1364--1372},
  year={2025}
}

@article{Adamw,
  title={Decoupled weight decay regularization},
  author={Loshchilov, Ilya and Hutter, Frank},
  journal={arXiv preprint arXiv:1711.05101},
  year={2017}
}

@article{uavtrack112,
  title={Onboard real-time aerial tracking with efficient Siamese anchor proposal network},
  author={Fu, Changhong and Cao, Ziang and Li, Yiming and Ye, Junjie and Feng, Chen},
  journal={IEEE Transactions on Geoscience and Remote Sensing},
  volume={60},
  pages={1--13},
  year={2021},
  publisher={IEEE}
}

@article{statrack,
  title={Fully Spiking Neural Networks with Target Awareness for Energy-Efficient UAV Tracking},
  author={Zhong, Pengzhi and Mo, Jiwei and Zeng, Dan and He, Feixiang and Li, Shuiwang},
  journal={arXiv preprint arXiv:2603.27493},
  year={2026}
}

@inproceedings{MAE,
  title={Masked Autoencoders Are Scalable Vision Learners},
  author={He, Kaiming and Chen, Xinlei and Xie, Saining and
          Li, Yanghao and Doll{\'a}r, Piotr and Girshick, Ross},
  booktitle={Proceedings of the IEEE/CVF Conference on
             Computer Vision and Pattern Recognition},
  pages={16000--16009},
  year={2022}
}

\end{document}